\documentclass[11pt,letterpaper]{mystyle}

\usepackage[table]{xcolor}
\usepackage{url}
\usepackage{xspace}
\usepackage{graphicx}
\usepackage{subcaption}
\usepackage{listings}
\usepackage{xcolor}
\usepackage{titletoc}
\usepackage{amsmath}
\usepackage[font={small}]{caption}
\usepackage[export]{adjustbox}
\usepackage{bm}
\usepackage{booktabs}
\usepackage{balance}
\usepackage{color}
\usepackage{dsfont}
\usepackage{esvect}
\usepackage{float}
\usepackage{graphicx}
\usepackage{pifont}
\usepackage{import}
\usepackage{mathtools}
\usepackage{multirow}
\usepackage{stackengine}
\usepackage{stmaryrd}
\usepackage{systeme}
\usepackage{soul}
\usepackage{tabularx}
\usepackage{url}
\usepackage{tikz}
\usetikzlibrary{tikzmark}
\usetikzlibrary{calc}
\usepackage[normalem]{ulem}
\usepackage{enumitem}
\usepackage{algorithm2e}
\usepackage{scalerel}
\usepackage{wrapfig}
\usepackage{bxcoloremoji}

\definecolor{codeblue}{rgb}{0.25,0.5,0.5}
\definecolor{nvgreen}{rgb}{0.92, 0.97, 0.85}

\definecolor{navyblue}{HTML}{0071BC}
\definecolor{hotpink}{HTML}{FF0080}

\definecolor{mygreen}{HTML}{FFD500}
\definecolor{myred}{rgb}{1, 0.9, 0.9}
\definecolor{mygray}{gray}{0.95}
\definecolor{mydarkblue}{rgb}{0,0.08,1}
\definecolor{mydarkred}{rgb}{0.8,0.02,0.02}
\definecolor{mydarkorange}{rgb}{0.40,0.2,0.02}
\definecolor{mypurple}{RGB}{111,0,255}
\definecolor{mygold}{rgb}{0.75,0.6,0.12}
\definecolor{mydarkgray}{rgb}{0.66, 0.66, 0.66}
\definecolor{mydarkgreen}{rgb}{0.02,0.6,0.02}
\definecolor{mygray}{gray}{0.9}
\definecolor{keynotegreen}{rgb}{0.04,0.52,0}
\definecolor{keynoteyellow}{rgb}{1,0.68,0}
\definecolor{LightCyan}{rgb}{0.88,1,1}
\definecolor{tabfirst}{rgb}{1, 0.7, 0.7}
\definecolor{tabsecond}{rgb}{1, 0.85, 0.7} 
\definecolor{tabthird}{rgb}{1, 1, 0.7} 
\definecolor{rbtred}{rgb}{255, 0, 0}

\newcommand{\myparagraph}[1]{\vspace{0.1mm}\noindent\textbf{#1}}

\SetKwComment{Comment}{\color{green!50!black}\# }{}

\SetKwProg{Function}{def}{:}{}

\SetKwProg{For}{for}{:}{}
\SetKwProg{If}{if}{:}{}

\def\method{SR-3D\xspace}
\def\bench{SR-3D-Bench\xspace}

\definecolor{citecolor}{HTML}{0071bc}
\usepackage{hyperref}[citecolor=citecolor]

\title{3D Aware Region Prompted Vision Language Model}

\author[1]{An-Chieh Cheng}
\author[1]{Yang Fu}
\author[3]{Yukang Chen}
\author[3]{Zhijian Liu}
\author[3]{Xiaolong Li}
\author[3]{Subhashree Radhakrishnan}
\author[2,3]{Song Han}
\author[3]{Yao Lu}
\author[3]{Jan Kautz}
\author[3]{Pavlo Molchanov}
\author[3,*]{Hongxu Yin}
\author[1,*]{Xiaolong Wang}
\author[3,*]{Sifei Liu}

\affiliation[1]{UC San Diego}
\affiliation[2]{MIT}
\affiliation[3]{NVIDIA}

\contribution[*]{Equal Advising}

\begin{document}
\abstract{
We present \textbf{S}patial \textbf{R}egion \textbf{3D} (\textit{SR-3D}) aware vision–language model that connects single-view 2D images and multi-view 3D data through a shared visual token space.
SR-3D supports flexible region prompting, allowing users to annotate regions with bounding boxes, segmentation masks on any frame, or directly in 3D, without the need for exhaustive multi-frame labeling.
We achieve this by enriching 2D visual features with 3D positional embeddings, which allows the 3D model to draw upon strong 2D priors for more accurate spatial reasoning across frames, even when objects of interest do not co-occur within the same view.
Extensive experiments on both general 2D vision language and specialized 3D spatial benchmarks demonstrate that SR-3D achieves state-of-the-art performance, underscoring its effectiveness for unifying 2D and 3D representation space on scene understanding. Moreover, we observe applicability to in-the-wild videos without sensory 3D inputs or ground-truth 3D annotations, where SR-3D accurately infers spatial relationships and metric measurements.
}
% \coloremojicode{1F3E0} \textbf{Project}: \href{https://tmp.github.io/}{https://tmp.github.io/}

% \coloremojicode{1F4DA} \textbf{Weights}: \href{https://huggingface.co/}{https://huggingface.co/}
% }

\metadata[\coloremojicode{1F3E0} \textbf{Project}]{\url{https://www.anjiecheng.me/sr3d}}
% \metadata[\coloremojicode{1F4DA} \textbf{Code}]{\url{https://huggingface.co/collections/}}    

\maketitle

\begin{figure*}[h]
\vspace{-0.1in}
\center
\includegraphics[width=\textwidth]
{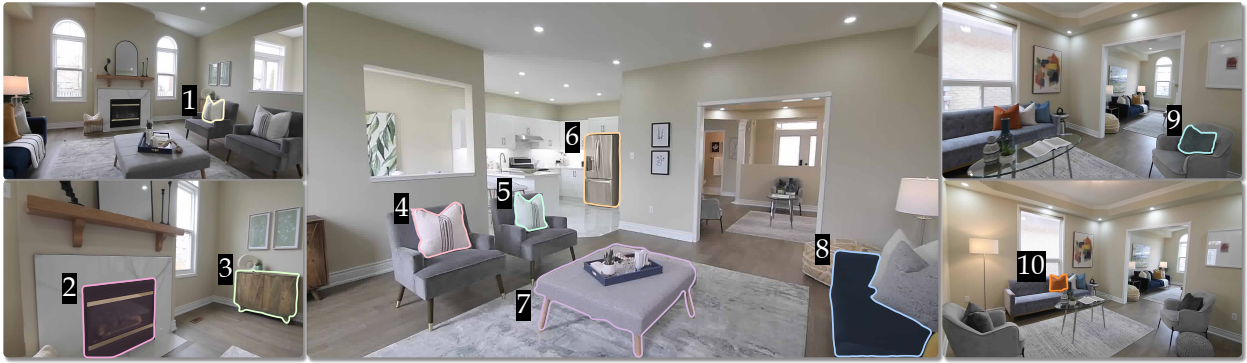}
\vspace{-0.25in}
\caption{\label{fig:teaser}From precise region-based distance estimation (\textit{left}), to intricate multi-view region query (\textit{middle}), and global cross-frame reasoning (\textit{right}), SR-3D delivers flexible and accurate spatial understanding to foundational Vision-Language Models. Notably, this video is obtained in the wild, \textbf{without sensory 3D inputs}, showcasing the remarkable generalization capability of our model.}
% \vspace{-1.2mm}
% Notably, this example is performed in the wild, \textbf{without camera parameters}, showcasing the remarkable generalization capability of our model.
\end{figure*}

\section{Introduction}
\label{sec:intro}
The rapid advancement of Vision Language Models (VLMs) ~\citep{openai2024gpt4o,liu2024visual,team2023gemini,wang2024qwen2,bai2025qwen2,nvila} has demonstrated strong capabilities in visual understanding~\citep{pratt2023does,huang2024lita} and language grounding~\citep{lv2023kosmos}. However, extending these strengths to 3D-aware spatial reasoning remains challenging. Foundational 2D VLMs excel at interpreting planar images, but generally lack mechanisms to capture complex 3D structural relationships. In contrast, most 3D VLMs~\citep{hong20233d,huang2023embodied,chatscene,xu2024pointllm,chen2024ll3da} operate in a fundamentally different representation space, making it difficult to leverage the prior knowledge from foundational 2D VLMs. Their performance is often hindered by limited 3D training data. Moreover, specifying spatial relationships solely through language can be cumbersome in cluttered scenes, e.g., multiple objects of the same category can coexist. A more direct way of specifying object instances is highly desirable.

To mitigate these challenges, recent efforts adopt multi-view images as a 3D representation that aligns seamlessly with the input space of foundational 2D VLMs~\citep{llava3d,video3dllm}. Unlike point clouds~\citep{huang2023embodied,chatscene,xu2024pointllm} which require extensive data collection and model alignment, a multi-view approach leverages strong 2D priors for 3D scene understanding. To specify object instances during reasoning, region prompts have proven effective in single-view VLMs~\citep{guo2024regiongpt,srgpt,yuan2023osprey,rasheed2023glamm}. However, extending region prompting to multi-view settings remains challenging. Specifically, an object may appear across different views with varying visibility, making comprehensive multi-frame or 3D bounding box annotation tedious and text-based queries imprecise. Ideally, a practical 3D-aware VLM should allow straightforward region annotations, such as marking a bounding box on a single frame, while still accurately reasoning about spatial relationships across the entire multi-view scene.

Thus, we introduce \method, a unified visual representation for 3D spatial understanding that leverages robust 2D foundational priors and supports flexible region prompting. In contrast to previous approaches that incorporate positional information only at 3D finetune stages~\cite{video3dllm}, or in different pathways~\cite{llava3d}, we directly integrate positional embeddings within the foundational VLM. Specifically, we estimate each input image's depth using an off-the-shelf depth estimator~\cite{depth_anything_v2} and transform this depth map into normalized 3D positional embeddings. For multi-view inputs representing a coherent scene, we further unify these positional embeddings into a common 3D coordinate space using either provided ground-truth camera poses or a point cloud estimator~\cite{dust3r,mast3r,cut3r} when only video inputs are available. Additionally, we incorporate region tokens directly into user prompts and train these region embeddings consistently at both the foundational single-view stages and the multi-view fine-tuning stage. Since the foundational VLM employs a dynamic tiling-based visual encoder~\cite{internvl,nvila}, we design a novel branch specifically compatible with this architecture to produce robust region embeddings.

The SR-3D architecture naturally supports flexible region annotation, enabling users to specify regions on any chosen frame. This practical capability arises from two key design choices: first, the consistent 3D positional embeddings in a canonical space enable the model to find coherent correspondences across frames; Second, the aligned embedding space from the foundational single-view stage naturally enables region embeddings to generalize effectively to the multi-frame contexts. As compelling evidence, our 2D-VLM trained exclusively on single-view data exhibits strong zero-shot spatial reasoning in 3D scenes, both with and without region prompts, despite never having been trained on multi-view data.

% (paragraph for experiment and contributions)
We conduct extensive evaluations across single-view and 3D multi-view settings, covering both region-level and global question-answering, each with general and spatial-related tasks. Our experiments demonstrate significant improvements in region-level performance. Specifically, our foundational 2D-VLM outperforms prior state-of-the-art methods by a large margin on region-level tasks, excelling in both recognition and spatial understanding. Additionally, we evaluate it on general VQA benchmarks and show that these improvements come without compromising overall VQA performance while also bringing benefits for general tasks that require spatial knowledge. For the 3D fine-tuned VLM, our model establishes new state-of-the-art results across general 3D question-answering, 3D video spatial understanding, and video region-level spatial tasks.  

Our contributions are as follows:

\begin{itemize}
\item We introduce \method, the first 3D-aware vision-language model that unifies representations for both single-view and multi-view tasks.
% \item We introduce \method, the first unified VLM capable of effective spatial reasoning across both 2D and 3D tasks. 
% enabled by a canonical positional embedding space shared between single- and multi-view scenarios. This unified design allows our model to seamlessly generalize spatial reasoning from 2D inputs to complex 3D scenes.
% \item We propose a novel dynamic tiling-based region extractor that captures robust region embeddings. This enables users to specify regions on any frame while ensuring spatial consistency across views through region embeddings trained in both single-view and multi-view settings.
\item We propose a dynamic tiling-based region extractor that handles high-resolution images and produces robust region embeddings. Our unified embedding space enables region representations trained on 2D images to generalize towards multi-view context.
\item  \method achieves state-of-the-art results in general 3D QA, video spatial reasoning, and region-based video tasks, demonstrating strong generalization and scalability.
\item We demonstrate real-world applications where our model effectively handles in-the-wild captured videos without 3D annotations (Figure~\ref{fig:teaser}), and can be flexibly prompted with region-level inputs.
%, accurately inferring spatial relationships and metric measurements.
\end{itemize}
\section{Related Work}
\label{sec:related-work}

\myparagraph{Region-level Vision-Language Models.}
% SIFEI: NEED TO CITE OMNIRGPT
% CAN BE SHRINKED WHEN SPACE limited
% BECAUSE the last paragraph is the most important
Region-level VLMs enhance fine-grained visual understanding by focusing on specific regions in images and videos. Early methods~\citep{peng2023kosmos,chen2023shikra,chen2023minigpt,wang2023cogvlm} represent regions as text using bounding box coordinates, making integration easy but relying on the language decoder for spatial reasoning. Others use visual markers like SoM~\citep{yang2023set}, which overlay numbers and masks but alter image appearance and require rule-based placement. Another approach maps region features into LLM tokens using RoI-aligned features~\citep{wang2023allseeing,wang2024allseeing_v2,wang2023visionllm,zhou2023regionblip,zhang2023gpt4roi,rasheed2023glamm,zhao2024dynrefer}, with RegionGPT~\citep{guo2024regiongpt} and Osprey~\citep{yuan2023osprey} refining this by pooling pixel-level mask features for flexible region shapes. However, they struggle with resolution and aspect ratio constraints. In the video domain, various representations~\citep{wang2024elysium,yu2024merlin,motionepic,ye2024x,omnirgpt} have been explored, but they mainly focus on tracking rather than multi-view spatial reasoning.

\myparagraph{Spatial Reasoning in Vision-Language Models.} Vision-language models have a strong visual understanding because they integrate the reasoning abilities of LLMs with powerful vision foundation models. Recently, there has been growing interest in equipping VLMs with spatial reasoning capabilities~\citep{chen2024spatialvlm,ma2024spatialpin,cai2024spatialbot,yuan2024robopoint,ma20243dsrbench,tang2024sparkle,song2024robospatial,xu2024llava,marsili2025visual,liu2025spatialcot,yang2025magma,liao2024reasoning}. While most previous work has focused on spatial understanding from 2D images, multi-view spatial reasoning remains less explored. Recently, VSI-Bench~\citep{yang2024thinking} was introduced as a testbed for evaluating models' 3D video-based spatial understanding. Our work extends this direction by proposing a unified 3D-aware architecture and representation that seamlessly supports both images and videos.

\myparagraph{3D Large Multimodal Models.}
Our work also relates to recent advancements in 3D LMMs~\citep{man2024situational,linghu2024multi,tang2024minigpt,hong20233d,fu2024scene,chen2024ll3da,huang2023embodied,wang2023chat}. Various 3D representations have been explored to integrate position information into LLMs. 3D-LLM~\citep{hong20233d} and Scene-LLM~\citep{fu2024scene} use multi-view images with object segmentation masks to construct pixel-aligned point representations, while LL3DA~\citep{chen2024ll3da} directly employs a point cloud encoder to extract 3D scene features. LEO~\citep{huang2023embodied} and Chat3D~\citep{wang2023chat} segment objects from the scene's point cloud and extract object features to represent the environment. These methods typically transform 3D scenes into voxel or point representations, but such approaches often limit the effectiveness of LLMs. Aligning these representations with LLMs requires vast amounts of data, which is challenging due to the scarcity of large-scale 3D datasets. Moreover, many of these methods rely on off-the-shelf 3d detection or segmentation models, which inherently constrain performance. 

The most closely related works to ours are LLaVA-3D~\citep{llava3d} and Video-3D-LLM~\citep{video3dllm}, which also incorporate 3D position-aware features into 2D vision-language models. However, LLaVA-3D processes 3D and 2D data through separate pathways, while Video-3D-LLM fine-tunes 3D video data on a pre-trained video VLM. Both approaches risk overfitting 3D position encodings to specific 3D tasks. In contrast, our method adopts a unified architecture and 3D representation space for both image and video data, enabling better alignment and improving generalization across spatial understanding tasks.

\begin{figure*}[!t]
    \centering
    % svila_fig2_v2.pdf
    \includegraphics[width=0.9\textwidth]{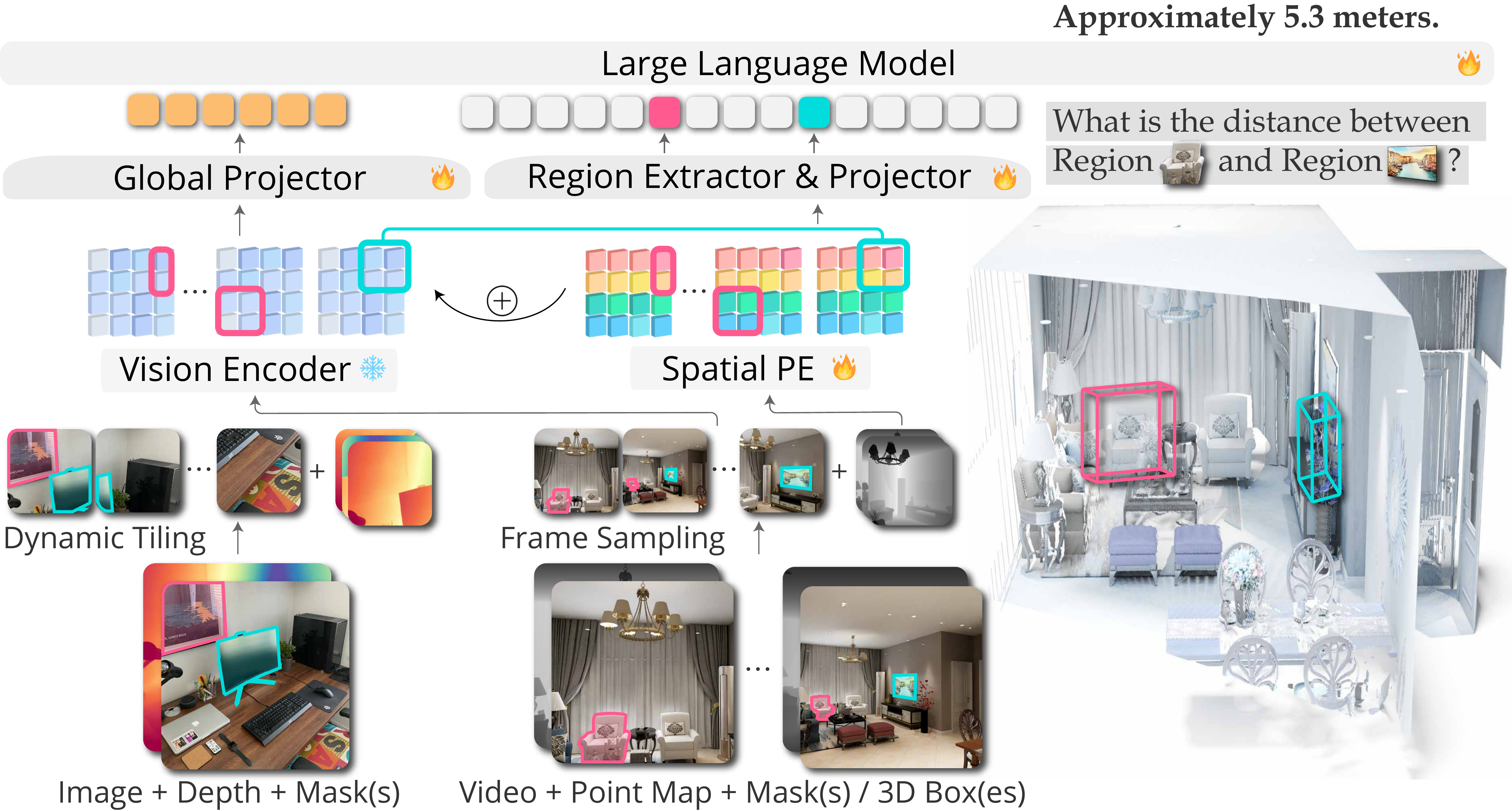}
    \vspace{-0.2cm}
    \caption{The SR-3D architecture. Given an image or multi-view input with optional region prompts (e.g., bounding boxes or masks), we encode them along with depth-derived positional embeddings using a tiling approach. Region tokens are extracted by stitching masked features, while 3D positional embeddings are mapped to a shared canonical space in the multi-view setting, as shown on the bottom right.}
    \label{fig:method}\vspace{-3mm}
\end{figure*}

\section{Methodology}
\label{sec:method}
We introduce a unified 3D-aware VLM architecture designed for both single-view and multi-view spatial understanding. Our approach leverages the strong priors of a foundational 2D model to infer spatial relationships across frames accurately. This is achieved by directly integrating 3D positional embeddings into the foundational 2D visual representations. To further enhance spatial grounding at the region level, we introduce a flexible and efficient module, the Dynamic Tiling-based Region Extractor, which operates seamlessly across both single- and multi-view inputs. As illustrated in Figure~\ref{fig:method}, our framework consists of a vision encoder, a 3D position encoding module, a region extractor, and an LLM backbone. In this section, we detail three key components: 
% (1) a canonical 3D positional representation for single- and multi-view images (Sec.\ref{sec:method:pe}), (2) the region extractor (Sec.\ref{sec:method:region}), and (3) the training paradigm (Sec.\ref{sec:method:train}), along with how our model operates during inference (Sec.~\ref{sec:method:inference}).
(1) a canonical 3D positional representation for single- and multi-view images (Sec.\ref{sec:method:pe}), (2) the region extractor (Sec.\ref{sec:method:region}), and (3) the training paradigm (Sec.\ref{sec:method:train}), along with how our model operates during inference (Sec.~\ref{sec:method:inference}).

\vspace{-2mm}
\subsection{Canonical 3D Positional Representation}
\label{sec:method:pe}
\vspace{-1mm}
A key idea of \method is the introduction of a canonical positional feature that is shared across both single-view and multi-view inputs. This unified representation allows us to leverage large-scale single-view image pretraining, while seamlessly transferring the learned spatial priors to multi-view scenarios.

\paragraph{Single-View Representation.}
We begin by pretraining our foundational VLM on large-scale 2D images to establish strong visual-language priors.  
Given a single-view image \( I \), we estimate its relative depth map \( D \) using DepthAnythingV2 \citep{depth_anything_v2}. We then compute a pixel-wise 3D position map in the camera coordinate system via back-projection, which is further canonicalized into a normalized world coordinate system. This canonicalization ensures that spatial information is expressed in a consistent and unified space, independent of camera pose.  

To inject spatial information into VLM, we encode the corresponding 3D position map into embeddings using a sinusoidal function followed by a learnable point-wise MLP. These embeddings are resized to align with the token dimensions and then added to their respective vision tokens. This fusion enriches visual representations with geometric awareness, enabling the model to better capture object placement and spatial relationships within the scene.  

\paragraph{Multi-View Representation.}
Building on the shared canonical space, we fine-tune the VLM with multi-view inputs to extend spatial reasoning beyond single images. We uniformly sample 32 frames from a video and resize both the images and their point maps to match the vision encoder’s input resolution. For multi-view training, we use ground-truth depth rather than estimated depth, performing back-projection and camera transformation to align the frames. The transformed point maps are normalized into the same canonical space as in the single-view setup, ensuring consistency in spatial representation. These processed frames and point maps act as the multi-view analog of the single-view tiles, enabling seamless integration of spatial and visual information across both training stages.

\subsection{Dynamic Tiling-based Region Extractor}
\label{sec:method:region}
\paragraph{Background: Dynamic Tiling-based Encoder.}
% The visual backbone produces a low-resolution feature map, which limits its ability to represent small-scale regions and objects when extracting region features. Prior works address this issue by employing a feature refinement module with deconvolution layers to upsample visual tokens. However, this refinement occurs after the vision encoder, meaning the features have already undergone resizing and potential distortion, which may restrict the effectiveness of the refinement process.
The visual backbone produces a low-resolution feature map, limiting its ability to represent small-scale regions and objects. To address this, we adopt the dynamic tiling mechanism employed in \citep{nvila} that enables high-resolution processing while maintaining spatial consistency. Instead of resizing entire images, we first determine the optimal aspect ratio by selecting the closest match from a predefined set (e.g., {1:1, 1:2, 2:1, 3:1, …, 2:6}), minimizing distortions. We then resize both the image and any corresponding point map accordingly and divide them into tiles of $448\times448$, matching the vision encoder's resolution. Each tile is encoded separately before being stitched back together, preserving local details without exceeding memory constraints. This tiling process is applied consistently across single-view and multi-view inputs, forming the basis for both our 3D positional embedding and region feature extraction strategies.

\vspace{-2mm}
\paragraph{Dynamic Region Extractor.}
Prior architectures without dynamic tiling rely on feature refinement modules with deconvolution layers to upsample visual tokens \citep{srgpt,guo2024regiongpt}, attempting to recover lost details. However, this refinement occurs after the vision encoder, meaning the features have already undergone resizing and potential distortion, which may limit its ability to fully recover fine details.

To address this, we introduce a \textit{tile-then-stitch} approach to extract region embeddings from high-resolution features. For single-view input, given a region of interest (RoI) represented by a binary mask, we apply the same dynamic tiling process used in the image pipeline to generate tiles of both the image and the mask. The tiled visual tokens and masks are then stitched back together at a higher resolution, followed by a mask-pooling operation to obtain the final mask feature.
This method offers two key advantages: (1) the extracted mask feature is derived from high-resolution features directly, reducing distortion and eliminating the need for post-refinement, and (2) our tile-then-stitch approach extends naturally to multi-view video inputs. In the multi-view setting, each frame is treated as a tile, allowing us to handle one or multiple masks per frame while maintaining spatial consistency across frames for the same RoI.
% \subsection{Spatial-related Data Curation}
% \AJ{2d/3d, region-level, and global spatial dataset}
% \label{sec:method:data}

\subsection{Training Paradigm}
\label{sec:method:train}
For the single-view VLM, we initialize the weights from a pre-trained 2D VLM (NVILA-Lite-8B~\cite{nvila}), keeping the vision encoder frozen while fine-tuning the 3D positional encoding module, projectors, and the LLM. We reuse the instruction fine-tuning dataset from the pre-trained VLM and blend it with region-prompted datasets \citep{guo2024regiongpt,srgpt} in this stage, resulting in a total data blend of approximately 7 million samples. Full dataset details are provided in the Supplementary Materials.

For the multi-view model, we fine-tune the single-view model using datasets such as ScanQA~\citep{azuma2022scanqa}, SQA3D~\citep{ma2024sqa3d}, and Scan2Cap~\citep{chen2021scan2cap}, as well as a newly curated EmbodiedScan~\citep{wang2024embodiedscan} dataset with region- and spatial-focused question-answer pairs. To enhance robustness and generalization, we apply various mask augmentations during multi-view training, including converting segmentation masks into bounding boxes and randomly dropping frames to simulate single-frame annotations. These strategies help the model learn to associate regions across frames while preserving spatial consistency. 

We note that, unlike prior work~\citep{llava3d} that employs separate pathways for single- and multi-view data, we adopt a unified pipeline where all data flows through the same model architecture. This ensures consistent processing of both single-view and multi-view inputs without distinction between spatial region prompts and global queries, allowing seamless integration of spatial reasoning at different levels.

\subsection{Inference}
\label{sec:method:inference}
% Our tile-and-stitch design provides a flexible mechanism for region-based inference. During training, we introduce random augmentations, such as converting masks into bounding boxes and dropping multi-view masks into single-view masks. As a result, during inference, our model can seamlessly handle different types of region inputs. For single-view inference, the model can accept either bounding boxes or segmentation masks as region inputs. In the multi-view setting, the model supports 3D bounding boxes, which are projected into multi-frame masks, or a single mask applied to a specific frame.
Our tile-and-stitch design enables flexible region-based inference. For single-view inputs, the model accepts bounding boxes or segmentation masks as region annotations. In multi-view scenarios, it supports a range of mask specifications: 3D bounding boxes that project into multi-frame masks, sparse-frame masks, or even a single-frame mask—reflecting our method’s ability to handle varying annotation densities while preserving spatial alignment.

For 3D input, although ground-truth depth maps were used during multi-view training, our approach remains highly adaptable due to the canonicalization of 3D positions into a normalized space. This allows us to replace ground-truth depth with point maps estimated from off-the-shelf models such as MAST3R \citep{mast3r} or CUT3R \citep{cut3r}. Our model offers a highly flexible and generalizable solution for spatial reasoning across diverse input modalities by maintaining a unified architecture that normalizes spatial information across different 3D sources.

\begin{table*}[t]
    \small
    \centering
    \setlength{\tabcolsep}{1pt}
    \scalebox{0.97}{
    {\fontsize{8pt}{9pt}\selectfont
    \begin{tabular}{lcccclclccccclccc}
    \toprule
    & \multicolumn{4}{c}{Spatial} & & \multicolumn{1}{c}{Math} & & \multicolumn{5}{c}{General Knowledge} & & \multicolumn{3}{c}{OCR-Related} \\
    \cmidrule(lr){2-6} \cmidrule(lr){7-7} \cmidrule(lr){9-13} \cmidrule(lr){15-17}
    Methods & BLINK$_S$ & SAT & EmbSpat & RWQA & & MathVista & & GQA& AI2D & MMMU$_p$ & SEED$_I$ & POPE & & Text$_{VQA}$ & Chart$_{QA}$ & Doc$_{VQA}$ \\
    \midrule
    % \color[HTML]{969696}SpatialRGPT-8B & & & & & \color[HTML]{969696}64.1 & & \color[HTML]{969696}74.5 & \color[HTML]{969696}85.5 & & \color[HTML]{969696}68.3 & & \\
    NVILA-Lite-8B & 79.7 & 62.6 & 68.9 & 65.6 & & 64.5 & & \bf65.3 & \bf91.0 & \bf25.1 & 76.3 & \bf88.1 & & \bf78.1 & \bf84.8 & \bf91.7 \\
    \rowcolor{mygreen!60}
    \textbf{\method-8B} & \phantom{$_\texttt{+4.2}$}\textbf{83.9}\textcolor{red}{$_\texttt{+4.2}$} & \phantom{$_\texttt{+1.4}$}\textbf{64.0}\textcolor{red}{$_\texttt{+1.4}$} & \phantom{$_\texttt{+3.6}$}\textbf{72.5}\textcolor{red}{$_\texttt{+3.6}$} & 
    \phantom{$_\texttt{+2.5}$}\textbf{68.1}\textcolor{red}{$_\texttt{+2.5}$} & & \bf65.4 & & 64.2 & 90.7 & 24.6 & \bf77.8 & 87.6 & & 77.3 & 83.9 & 91.0 \\
    \bottomrule
    \end{tabular}}}
    \vspace{-0.2cm}
    \caption{Comparison of \method and base model~\citep{nvila} performance on general image VQA benchmarks.}
    \label{tab:2d_general}
    \vspace{-0.15cm}
\end{table*}

\begin{figure}[!t]
    \centering
    \includegraphics[width=1.0\linewidth]{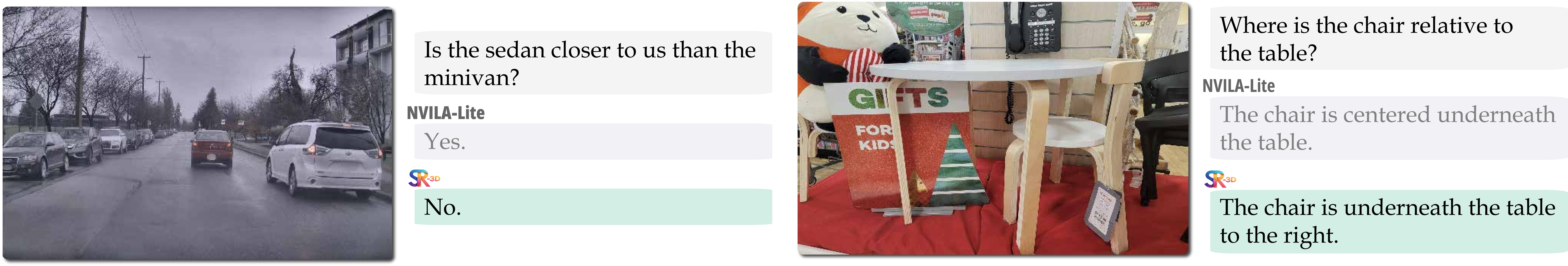}
    \vspace{-0.5cm}
    \caption{RealWorldQA results. \method shows stronger spatial understanding of physical environments compared to the base model. We omit the answer choices for clarity in visualization.}
    \label{fig:rwqa}
\end{figure}

\section{Experiments}
\label{sec:exp}
We first evaluate \method on 2D benchmarks (Section~\ref{sec:exp:2d}) to verify whether the introduced positional features improve performance while preserving the generalization of the base single-view model. We then evaluate the multi-view model on 3D benchmarks in Section~\ref{sec:exp:3d}. We further show ablation studies in Section~\ref{sec:exp:abl} to analyze the role of pretraining and 3D positional encoding. Finally, we demonstrate that our method can be seamlessly integrated with off-the-shelf 3D geometry foundation models as an application (Section~\ref{sec:exp:cut3r}).

\subsection{Evaluation on 2D Benchmarks}
\label{sec:exp:2d}
\myparagraph{Region-level Question Answering.}
We evaluate our model’s object classification performance on the COCO-2017~\citep{lin2014microsoft} dataset using mean Average Precision (mAP) and classification accuracy as metrics. Following prior work on region-level recognition~\citep{zhong2022regionclip,guo2024regiongpt,srgpt}, we rely on ground-truth boxes for positional information and augment the general prompt with task-specific instructions. As reported in Table~\ref{tab:2d-coco}, \method attains an mAP of 78.0 and an accuracy of 88.6\%, demonstrating strong region-level recognition and validating the effectiveness of our region extractor. Compared with SpatialRGPT~\citep{srgpt}, which is trained on the same region-level data, our model achieves significant gains, largely attributable to the dynamic tiling extractor that provides higher-fidelity regional masks. For reference, we also include DynRefer’s RoIAlign (448 variant)~\citep{zhao2024dynrefer} as a baseline at the same resolution. Importantly, their proposed strategies are complementary to our approach.

\begin{table*}[!t]
\small
\centering
\setlength{\tabcolsep}{10pt}

\begin{minipage}{0.48\textwidth}
\centering
\setlength{\tabcolsep}{2.5pt}
\scalebox{0.97}{
{\fontsize{8pt}{9pt}\selectfont
\begin{tabular}{lc}
    \toprule
    Methods & Acc. (\%) \\ \midrule
    \color[HTML]{969696}Human & \color[HTML]{969696}98.3 \\
    \midrule
    \rowcolor{navyblue!5}
    \multicolumn{1}{l}{\textbf{\emph{Proprietary Models (API)}}} & \\
    Qwen-VL-Max~\citep{bai2023qwen} & 58.9 \\
    Gemini Pro~\citep{team2023gemini} & 50.0 \\
    Claude 3 OPUS~\citep{claude3} & 57.3 \\
    GPT-4V-\textit{preview}~\citep{openai2024gpt4o} & 58.9 \\
    GPT-4V-\textit{Turbo}~\citep{openai2024gpt4o} & 66.9 \\
    GPT-4o~\citep{openai2024gpt4o} & 64.5 \\
    \midrule
    \rowcolor{navyblue!5}
    \multicolumn{1}{l}{\textbf{\emph{Open-source Models}}} & \\
    InstructBLIP-13B~\citep{dai2024instructblip} & 50.0 \\
    Yi-VL-34B~\citep{young2024yi} & 53.2 \\
    LLaVA-v1.5-13B-xtuner~\citep{2023xtuner} & 54.0 \\
    LLaVA-v1.6-34B~\citep{llavanext} & 64.5 \\
    MiniGPT-4-v2-7B~\citep{chen2023minigpt} & 49.2 \\
    InstructBLIP-7B~\citep{dai2024instructblip} & 50.8 \\
    LLaVA-v1.5-7B-xtuner~\citep{2023xtuner} & 50.8 \\
    CogVLM-7B~\citep{wang2023cogvlm} & 50.8 \\
    LLaVA-v1.5-7B~\citep{liu2024improved} & 51.6 \\
    LLaVA-InternLM2-7B~\citep{cai2024internlm2} & 52.4 \\
    SpatialRGPT-8B~\citep{srgpt} & 87.9 \\
    \rowcolor{mygreen!60}
    \textbf{\method-8B} & \textbf{90.3} \\
    \bottomrule
\end{tabular}}}
\caption{Results on BLINK$_\texttt{Depth}$. We follow SpatialRGPT~\cite{srgpt}’s protocol to test whether a 3D-injected VLM effectively leverages auxiliary spatial information.}
\label{tab:2d_blink}
\end{minipage}
\hfill
\begin{minipage}{0.48\textwidth}
\centering
\setlength{\tabcolsep}{2.5pt}
\scalebox{0.97}{
{\fontsize{8pt}{9pt}\selectfont
\begin{tabular}{lcc}
    \toprule
    Methods & mAP ($\uparrow$) & Acc. (\%) \\ \midrule
    CLIP~\citep{radford2021learning} & 58.9 & - \\
    RegionCLIP~\citep{zhong2022regionclip} & 58.3 & - \\
    \midrule
    LLaVA-7B~\citep{liu2024visual} & - & 40.0 \\
    Shikra-7B~\citep{chen2023shikra} & - & 53.9 \\
    GPT4RoI-7B~\citep{zhang2023gpt4roi} & - & 64.0 \\
    PVIT-7B~\citep{chen2023position} & - & 64.5 \\
    ASM-7B~\citep{wang2023all} & 69.3 & - \\
    RegionGPT-7B~\citep{guo2024regiongpt} & 70.0 & 80.6 \\
    DynRefer~\citep{zhao2024dynrefer} & - & 81.2 \\
    SpatialRGPT-8B~\citep{srgpt} & 72.9 & 82.9 \\
    \rowcolor{mygreen!60}
    \textbf{\method-8B} & \textbf{78.0} & \textbf{88.6} \\
    \bottomrule
\end{tabular}}}
\caption{Region-level classification results on COCO-2017 val set with ground-truth boxes, following RegionCLIP~\citep{zhong2022regionclip} and RegionGPT~\citep{guo2024regiongpt}.}
\label{tab:2d-coco}
\end{minipage}

\vspace{-3mm}
\end{table*}

We further evaluate \method on the BLINK$_\texttt{Depth}$ benchmark~\citep{fu2024blink} using the region-prompts as in SpatialRGPT~\citep{srgpt}, which tests point-level depth understanding in VLMs. BLINK$_\texttt{Depth}$ is a challenging task that requires both spatial and regional awareness. We report results in Table~\ref{tab:2d_blink} showing that \method outperforms current state-of-the-art SpatialRGPT~\citep{srgpt}, achieving 90\% accuracy. These results highlight that our approach excels in region extraction and effectively utilizes the provided 3D-aware input.

\begin{table*}[t]
    \small
    \centering
    \setlength{\tabcolsep}{4.75pt}
    \scalebox{0.97}{
    {\fontsize{8pt}{9pt}\selectfont
    \begin{tabular}{lcccclccccclc}
    \toprule
    & \multicolumn{4}{c}{Scan2Cap} & & \multicolumn{5}{c}{ScanQA} & & \multicolumn{1}{c}{SQA3D} \\
    \cmidrule(lr){2-5} \cmidrule(lr){7-11} \cmidrule(lr){13-13}
   Methods & B-4 $\uparrow$ & Rouge $\uparrow$ & Cider $\uparrow$ & Meteor $\uparrow$ & & B-4 $\uparrow$ & Rouge $\uparrow$ & Cider $\uparrow$ & Meteor $\uparrow$ & EM $\uparrow$ & & EM $\uparrow$ \\
    \midrule
\rowcolor{navyblue!5}  
    \multicolumn{13}{l}{\bf\emph{Task-specific Specialist}}\\
    VoteNet+MCAN~\citep{yu2019deep} & - & - & - & - & & 6.2 & 29.8 & 54.7 & 11.4 & 17.3 & & - \\
    ScanRefer+MCAN~\citep{yu2019deep} & - & - & - & - & & 7.9 & 30.0 & 55.4 & 11.5 & 18.6 & & - \\
    ScanQA~\citep{azuma2022scanqa}  & - & - & - & - & & 10.1 & 33.3 & 64.9 & 13.1 & 21.0 & & - \\
    3D-VisTA~\citep{zhu20233d} & 34.0 & 54.3 & 66.9 & 27.1 & & 10.4 & 35.7 & 69.6 & 13.9 & 22.4 & & - \\
    \midrule
\rowcolor{navyblue!5}  
    \multicolumn{13}{l}{\bf\emph{2D Large Multi-modal Models}}\\
    Oryx-34B~\citep{liu2024oryx} & - & - & - & - & & - & 37.3 & 72.3 & 15.0 & - & & - \\
    NaviLLM~\citep{zheng2024towards} & - & - & - & - & & 12.0 & 38.4 & 75.9 & 15.4 & 23.0 & & - \\
    LLaVA-Video-7B\textsuperscript{\dag}~\citep{zhang2024video} & - & - & - & - & & 3.1 & 44.6 & 88.7 & 17.7 & - & & - \\
    NaVILA~\citep{cheng2024navila} & - & - & - & - & & 16.9 & 49.3 & 102.7 & 20.1 & 28.6 & & - \\
    \midrule
\rowcolor{navyblue!5}    
    \multicolumn{13}{l}{\bf\textit{3D Large Multi-modal Models}}\\
    3D-LLM$_{(flamingo)}$~\citep{hong20233d}  & - & - & - & - & & 7.2 & 32.3 & 59.2 & 12.2 & 20.4 & & - \\
    3D-LLM$_{(BLIP2-flant5)}$~\citep{hong20233d}  & - & - & - & - & & 12.0 & 35.7 & 69.4 & 14.5 & 20.5 & & - \\
    LL3DA~\citep{chen2024ll3da} & 36.8 & 55.1 & 65.2 & 26.0 & & 13.5 & 37.3 & 76.8 & 15.9 & - & & - \\
    Chat-3Dv2~\citep{wang2023chat} & - & - & - & - & & 14.0 & - & 87.6 & - & - & & 54.7 \\
    LEO~\citep{huang2023embodied} & 36.9 & 57.8 & 68.4 & 27.7 & & 13.2 & 49.2 & 101.4 & 20.0 & 24.5 & & 50.0 \\
    Scene-LLM~\citep{fu2024scene} & - & - & - & - & & 12.0 & 40.0 & 80.0 & 16.6 & 27.2 & & 54.2 \\
    ChatScene~\citep{chatscene} & 36.3 & 58.1 & 77.2 & 28.0 & & 14.3 & 41.6 & 87.7 & 18.0 & 21.6 & & 54.6 \\
    LLaVA-3D~\citep{llava3d} & 41.1 & 63.4 & 79.2 & 30.2 & & 14.5 & 50.1 & 91.7 & 20.7 & 27.0 & & 55.6 \\
    Video-3D LLM~\citep{video3dllm} & 42.4 & 62.3 & 83.8 & 28.9 & & 16.2 & 49.0 & 102.1 & 19.8 & 30.1 & & 58.6 \\
    \rowcolor{mygreen!60}
    % \textbf{\method-8B} & \bf44.3 & \bf66.9 & \bf95.6 & \bf31.2 & & \bf18.7 & \bf50.5 & \bf107.2 & \bf20.8 & \underline{29.5} & & \bf60.5 \\
    \textbf{\method-8B} & \bf44.7 & \bf67.3 & \bf97.9 & \bf31.5 & & \bf18.1 & \bf51.2 & \bf109.3 & \bf21.2 & \bf30.4 & & \bf62.2 \\
    \bottomrule
    \end{tabular}}}
    \vspace{-0.2cm}
    \caption{Evaluation of spatial scene understanding performance on the Scan2Cap, ScanQA, and SQA3D benchmarks. \textsuperscript{\dag} indicates methods evaluated in a zero-shot setting. \method achieves state-of-the-art results across all metrics.}
    \label{tab:3d_general}
    % \vspace{-5mm}
\end{table*}

\myparagraph{General Question Answering.}
We investigate two key questions: (1) Does incorporating 3D positional information affect general vision-language understanding capabilities? (2) Can it improve performance on spatial-related tasks? To answer these, we evaluate our model on general VLM benchmarks covering Spatial~\citep{xai2024grok15,ray2024sat,du2024embspatial,fu2024blink}, Math~\citep{lu2024mathvista}, General Understanding~\citep{hudson2019gqa,kembhavi2016diagram,yue2024mmmu,li2024seed,li2023evaluating}, and OCR-related~\citep{singh2019towards,masry2022chartqa,mathew2021docvqa} tasks. As shown in Table~\ref{tab:2d_general}, compared to the base model NVILA-Lite-8B~\citep{nvila}, our model maintains comparable performance in math, general understanding, and OCR-related tasks, confirming that integrating 3D positional information does not degrade overall vision-language capabilities. Additionally, our method improves performance on the spatial understanding benchmark RealWorldQA~\citep{xai2024grok15}. We also provide qualitative examples from RealWorldQA in Figure~\ref{fig:rwqa}, showcasing cases where NVILA-Lite fails while \method succeeds in spatial reasoning tasks. These results demonstrate that our 3D-aware VLM enhances spatial reasoning while preserving general vision-language capabilities.

\subsection{Evaluation on 3D Benchmarks}
\label{sec:exp:3d}
\vspace{-1mm}
\myparagraph{General 3D Question Answering.}
We report results on three classic 3D vision-language understanding tasks: 3D dense captioning on Scan2Cap~\citep{chen2021scan2cap}, ScanQA~\citep{azuma2022scanqa}, and SQA3D~\citep{ma2024sqa3d}. Our evaluation metrics include conventional scores (e.g., CIDEr, BLEU, METEOR, ROUGE) as well as exact-match (EM) accuracy. Following prior work, we assume that input scenes may lack 3D object mask annotations during inference and use off-the-shelf models to generate proposals. However, unlike previous approaches, we leverage 2D segmentation models to generate 2D object proposals instead.
We compare NaVILA against strong baselines, including task-specific specialist models for each benchmark and leading methods from both 2D and 3D large multimodal models (LMMs). NaVILA significantly outperforms state-of-the-art single-task and task-specific fine-tuned models on 3D dense captioning and 3D QA tasks. 

\begin{table*}[t]
    \small
    \centering
    \setlength{\tabcolsep}{5.5pt}
    \scalebox{0.97}{
    {\fontsize{8pt}{9pt}\selectfont
    \begin{tabular}{lcccccccccc}

    & \rotatebox{55}{Wide/Thin} & \rotatebox{55}{Tall/Short} & \rotatebox{55}{Big/Small} & \rotatebox{55}{Multi. Simple} & \rotatebox{55}{Multi. Complex} & \rotatebox{55}{Avg.} & \rotatebox{55}{Width} & \rotatebox{55}{Height} & \rotatebox{55}{Distance} & \rotatebox{55}{Avg.} \\

    Methods & \multicolumn{6}{c}{\cellcolor{yellow!10}Qualitative} & \multicolumn{4}{c}{\cellcolor{orange!10}Quantitative} \\

    \midrule 
    \rowcolor{navyblue!5}
    \multicolumn{1}{l}{\textcolor{black}{\bf\emph{Blind LLMs w/ Language Referral}}} & & & & & & & & & &\\
    GPT-4o~\citep{openai2024gpt4o} & 64.8 & 64.5 & 64.0 & 47.8 & 41.4 & 56.5 & 70.5 & 70.6 & 50.4 & 63.8 \\
    \midrule 
    \rowcolor{navyblue!5}
    \multicolumn{1}{l}{\textcolor{black}{\bf\emph{VLMs w/ Language Referral}}} & & & & & & & & & & \\
    GPT-4o~\citep{openai2024gpt4o} & 52.1 & 54.1 & 57.5 & 62.4 & 42.4 & 53.7 & 72.4 & 72.8 & 55.8 & 67.0 \\
    NVILA-Video-8B~\citep{nvila} & 48.8 & 38.9 & 53.7 & 52.1 & 36.0 & 45.9 & 59.2 & 54.3 & 6.6 & 40.0 \\
    \midrule 
    \rowcolor{navyblue!5}
    \multicolumn{1}{l}{\textcolor{black}{\bf\emph{Region VLMs}}} & & & & & & & & & & \\
    GPT-4o~\citep{openai2024gpt4o}+SoM & 46.1 & 39.9 & 39.3 & 52.1 & 43.2 & 44.1 & 52.4 &  47.8 & 40.0 & 46.7 \\
    NVILA-Video-8B~\citep{nvila}+SoM & 49.3 & 40.0 & 53.7 & 52.1 & 40.4 & 47.1 & 59.3 & 54.1 & 6.6 & 40.0 \\
    \rowcolor{mygreen!60}
    % \method-8B-2D & 58.5 & 71.4 & 79.7 & - & - & - & 57.1 & 68.5 & 68.5 & - \\
    \rowcolor{mygreen!60}
    % \textbf{\method-8B} & \bf76.3 & \bf81.8 & \bf80.5 & \bf81.2 & \bf73.8 & \bf78.7 & \bf87.5 & \bf87.1 & \bf70.3 & \bf81.6 \\
    \textbf{\method-8B} & \bf76.3 & \bf83.1 & \bf81.8 & \bf80.3 & \bf76.0 & \bf79.5 & \bf87.7 & \bf87.3 & \bf74.8 & \bf83.3 \\
    \bottomrule
    \end{tabular}}}
    \vspace{-0.2cm}
    \caption{Evaluation of region-level spatial scene understanding on the \bench.}
    % \vspace{-0.5cm}
    \label{tab:3d_region}
\end{table*}
\begin{figure*}[ht]
    \centering
    \includegraphics[width=1.0\linewidth]{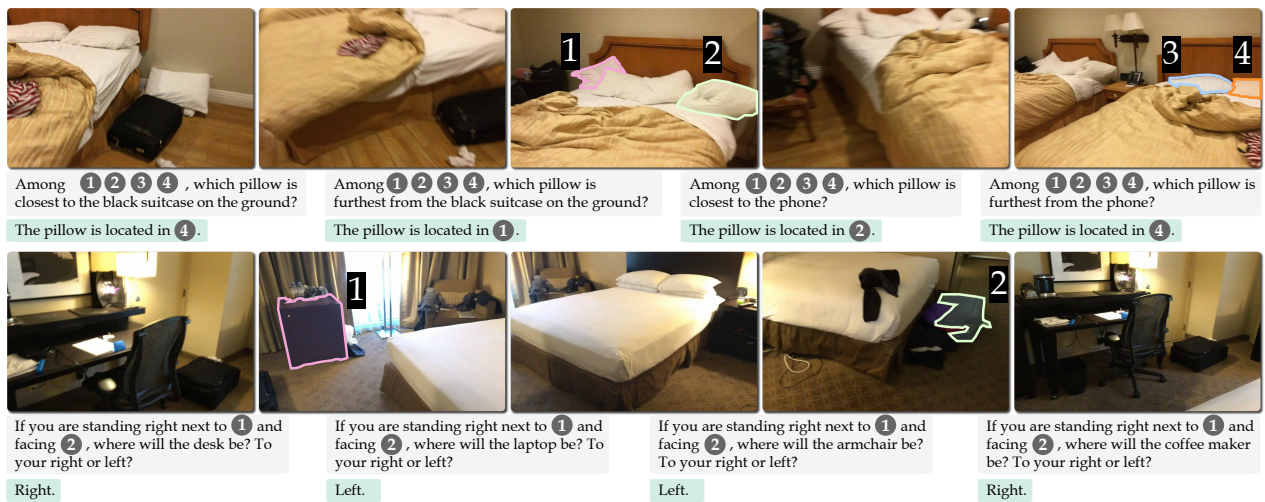}
    \caption{\method results on region-level multi-view spatial understanding. We show extreme cases where the same region prompts are used across samples but with different target objects. \method answers all queries correctly, showing strong evidence that it truly understands 3D spatial relationships.}
    \label{fig:region3d}
    \vspace{-0.4cm}
\end{figure*}

% \begin{figure*}[htbp]
%     \centering
%     \includegraphics[width=1.0\linewidth]{figures/results/vsi_qual.pdf}
%     \caption{SVILA-Bench Results.}
%     \label{fig:svilabench}
% \end{figure*}

\vspace{-1mm}
\subsection{Video Spatial Intelligence.}
\vspace{-1mm}
\myparagraph{Region-level Spatial QA.}
Currently, no video benchmarks specifically focus on region-level spatial understanding. Without explicit region information, spatial understanding can become ambiguous, especially when multiple identical objects are present or when referring to a specific area in a scene that is difficult to describe precisely using language alone. To address this, we propose \bench, a region-level spatial benchmark curated from ScanNet~\citep{dai2017scannet}, ARKitScenes~\citep{baruch2021arkitscenes}, and Matterport\citep{chang2017matterport3d} video scan datasets with 3D ground truth. Specifically, we utilize preprocessed oriented bounding box annotations from EmbodiedScan~\citep{wang2024embodiedscan}, where each object is axis-aligned within a canonicalized geodetic coordinate system. This alignment ensures that the bounding box dimensions accurately represent the true width, length, and height. Using these bounding boxes, we construct a conversational benchmark that includes both qualitative and quantitative question-answering tasks. The qualitative QA consists of choice-based, predicate-based, and multiple-choice questions, while the quantitative QA focuses on measuring object width, height, and distance. We generate these QA pairs using template-based conversation generation and allow the VLM to generate free-form language. For qualitative QA evaluation, we use GPT-4o~\citep{openai2024gpt4o} as an evaluator and report the accuracy, while for quantitative QA, we measure the success rate by thresholding the maximum ratio between estimation and the ground truth value.

We report three types of baseline models: (1) Blind LLMs, which answer questions using only the provided text without visual input. To improve this, we replace the mask prompt with the object class for each question. This serves as a baseline to measure how much video spatial reasoning can come from general world knowledge alone. We use GPT-4o as the representative, as it is one of the most advanced models for general knowledge. (2) VLMs with Language Referral, which have access to visual content, allowing them to potentially perform better than blind LLMs. We use state-of-the-art vision-language models GPT-4o~\citep{openai2024gpt4o} and NVILA-Video~\citep{nvila} as baselines in this category. (3) Region-aware Video VLMs. These models process specific image regions without relying on text descriptions or object class information. We equip GPT-4o and NVILA-Video with Set of Marks (SoM) for region-based reasoning. Note that while \citep{qiu2024artemis} and ~\citep{wang2024elysium} are also region-level video VLMs, they are excluded from comparisons as they cannot handle multi-object input or lack support for multi-frame prompts.

We present results in Table~\ref{tab:3d_region}. The findings suggest that both Blind LLMs and VLMs with Language Referral perform reasonably well on quantitative tasks, such as estimating object width, due to their general world knowledge. However, region-level VLMs equipped with SoM struggle, likely because the models find it challenging to track the set of marks across frames. Overall, our method outperforms all baselines across all categories.

\begin{table}[t!]
    % \vspace{-0.1cm}
    \centering
    \fontsize{8pt}{9pt}\selectfont
    \setlength\tabcolsep{6.5pt} 
    \scalebox{0.97}{
    \begin{tabular}{lccccccc}
    & 
    \rotatebox{55}{Abs. Dist.} &
    \rotatebox{55}{Obj. Size} & 
    \rotatebox{55}{Room Size} &
    \rotatebox{55}{Rel. Dist.} &
    \rotatebox{55}{Rel. Dir.} \\
    Methods & \multicolumn{3}{c}{\cellcolor{orange!10}Quantitative} & \multicolumn{2}{c}{\cellcolor{yellow!10}Qualitative} \\
    \midrule
    % \rowcolor{navyblue!5}
    % \multicolumn{1}{l}{\textcolor{black}{\bf\emph{Baseline}}} & & & & & \\
    \color[HTML]{969696}Random & \color[HTML]{969696}- & \color[HTML]{969696}- & \color[HTML]{969696}- & \color[HTML]{969696}25.0 & \color[HTML]{969696}36.1 \\
    \color[HTML]{969696}Human Level\textsuperscript{\dag} & \color[HTML]{969696}47.0 & \color[HTML]{969696}60.4 & \color[HTML]{969696}45.9 & \color[HTML]{969696}94.7 & \color[HTML]{969696}95.8 \\
    \midrule
    \rowcolor{navyblue!5}
    \multicolumn{1}{l}{\textcolor{black}{\bf\emph{Proprietary Models (API)}}} & & & & & \\
    GPT-4o~\cite{openai2024gpt4o} & 5.3 & 43.8 & 38.2 & 37.0 & 41.3 \\
    Gemini-1.5 Flash~\cite{team2024gemini} & 30.8 & 53.5 & {54.4} & 37.7 & 41.0 \\
    Gemini-1.5 Pro~\cite{team2024gemini} & {30.9} & {64.1} & 43.6 & {51.3} & {46.3} \\
    \midrule
    \rowcolor{navyblue!5}
    \multicolumn{1}{l}{\textcolor{black}{\bf\emph{Open-source Models}}} & & & & & \\
    InternVL2-2B~\cite{chen2024internvl} & 24.9 & 22.0 & 35.0 & 33.8 & {44.2} \\
    InternVL2-8B~\cite{chen2024internvl} & {28.7} & 48.2 & {39.8} & 36.7 & 30.7 \\
    InternVL2-40B~\cite{chen2024internvl} & 26.9 & 46.5 & 31.8 & 42.1 & 32.2 \\
    LongVILA-8B~\cite{xue2024longvila} & 9.1 & 16.7 & 0.0 & 29.6 & 30.7 \\
    VILA-1.5-8B~\cite{lin2024vila} & 21.8 & 50.3 & 18.8 & 32.1 & 34.8 \\
    VILA-1.5-40B~\cite{lin2024vila} & 24.8 & 48.7 & 22.7 & 40.5 & 25.7 \\
    LongVA-7B~\cite{zhang2024long} & 16.6 & 38.9 & 22.2 & 33.1 & 43.3 \\
    LLaVA-NeXT-Video-7B\cite{llavanext} & 14.0 & 47.8 & 24.2 & {43.5} & 42.4 \\
    LLaVA-NeXT-Video-72B~\cite{llavanext} & {22.8} & 57.4 & 35.3 & 42.4 & 36.7 \\
    LLaVA-OneVision-0.5B~\cite{li2025llavaonevision} & 28.4 & 15.4 & 28.3 & 28.9 & 36.9 \\
    LLaVA-OneVision-7B~\cite{li2025llavaonevision} & 20.2 & 47.4 & 12.3 & 42.5 & 35.2 \\
    LLaVA-OneVision-72B~\cite{li2025llavaonevision} & 23.9 & {57.6} & 37.5 & 42.5 & 39.9 \\
    \rowcolor{mygreen!60}
    \textbf{\method-8B} & \bf52.8 & \bf75.5  & \bf41.9 & \bf57.3 & \bf82.3 \\
    % \textbf{\method-8B} & \bf49.2 & \bf76.6 & \bf41.0 & \bf52.0 & \bf50.3 \\
    \bottomrule
    \end{tabular}
} 
    \vspace{-0.2cm}
    \caption{Results on multi-view global spatial scene understanding evaluated on VSI-Bench~\citep{yang2024thinking}. \textsuperscript{\dag} indicates methods tested on the \texttt{Tiny} subset. \method achieves strong performance on the relative direction task, providing clear evidence that the model effectively leverages the 3D positional encoding.}
    \label{tab:3d_vsibench}
    \vspace{-0.2cm}
\end{table}

\begin{figure*}[t]
    \centering
    \includegraphics[width=1.0\linewidth]{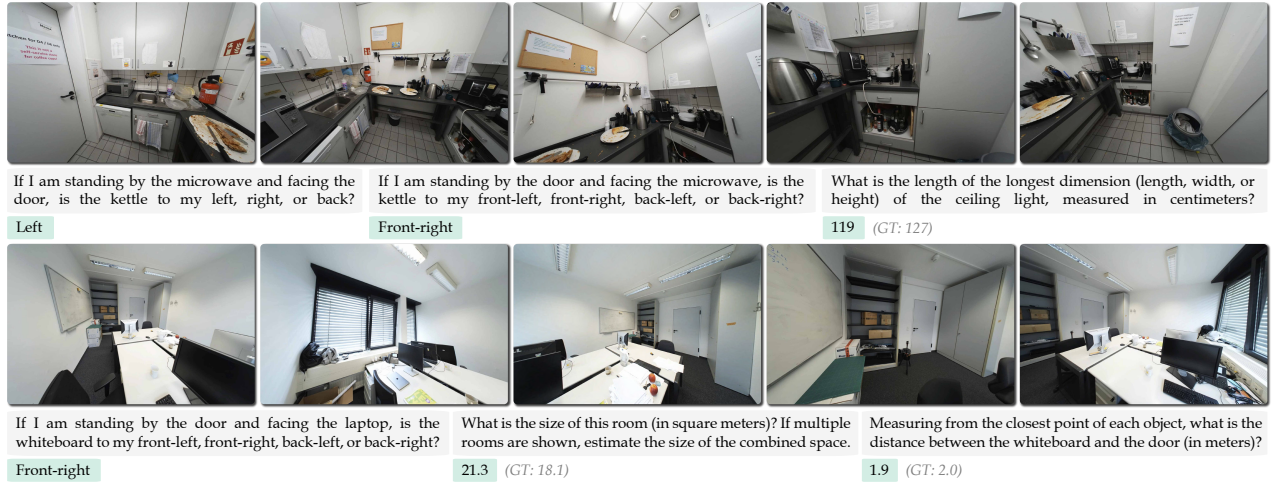}
    % \vspace{-0.3cm}
    \caption{VSI-Bench results. We highlight \method’s outputs and include ground-truth values for numerical answers. The results show that \method answers spatial questions correctly even without region prompts.}
    \label{fig:vsibench}
    % \vspace{-0.4cm}
\end{figure*}

\myparagraph{Global Spatial QA.}
We also report results on global spatial understanding using VSI-Bench~\citep{yang2024thinking}, a recently proposed benchmark that quantitatively evaluates the visual-spatial intelligence of VLMs based on egocentric videos. In our evaluation, we exclude categories that are less relevant to spatial reasoning, such as appearance order, which is more about temporal understanding. We use accuracy as the evaluation metric for qualitative questions and Mean Relative Accuracy (MRA) for quantitative questions. As shown in Table~\ref{tab:3d_vsibench}, \method outperforms all open-source models and performs comparably, if not better, than API-based models.

% \begin{table}[t]
%     \small
%     \centering
%     \setlength{\tabcolsep}{7.6pt}
%     \scalebox{0.97}{
%     {\fontsize{8pt}{9pt}\selectfont
%     \begin{tabular}{lcccc}
%     \toprule
%     Methods & Tall/Short & Big/Small & Height & Distance \\
%     \midrule 
%     \method-8B-2D & \underline{71.4} & \underline{79.7} & \underline{68.5} & \underline{68.5} \\
%     \method-8B & 83.1 & 81.8 & 87.3 & 74.8 \\
%     \bottomrule
%     \end{tabular}}}
%     \vspace{-0.2cm}
%     \caption{Zero-shot evaluation of our 2D-trained VLM on \bench, testing whether the model's representations are truly aligned. Our 2D model achieves reasonable accuracy without explicit 3D supervision.}
%     \label{tab:abl_zeroshot}\vspace{-0.35cm}
% \end{table}

\begin{table}[!t]
    \scriptsize
    \centering
    \setlength{\tabcolsep}{7.5pt} 
    \scalebox{0.97}{
    \begin{tabular}{@{}l c cccc@{}}
    \toprule
     & 2D Pre-train & 3D Tall/Short & 3D Big/Small & 3D Height & 3D Distance \\
    \midrule 
    \rowcolor{navyblue!5}
    \multicolumn{6}{@{}l}{\bf\emph{Zero-shot 2D Models}}\\
    Base Model & & \phantom{$_\texttt{-31.4}$}40.0\textcolor{red}{$_\texttt{-31.4}$} & \phantom{$_\texttt{-26.0}$}53.7\textcolor{red}{$_\texttt{-26.0}$} & \phantom{$_\texttt{-14.4}$}54.1\textcolor{red}{$_\texttt{-14.4}$} & \phantom{$_\texttt{-61.9}$}6.6\textcolor{red}{$_\texttt{-61.9}$} \\
    \rowcolor{mygreen!60}
    \bf\method-2D & \checkmark & \bf{71.4} & \bf{79.7} & \bf{68.5} & \bf{68.5} \\
    \midrule 
    \rowcolor{navyblue!5}
    \multicolumn{6}{@{}l}{\bf\emph{Finetuned 3D Models}}\\ % Spans all 6 columns
    \method & & \phantom{$_\texttt{-0.0}$}83.1\textcolor{red}{$_\texttt{0.0}$} & \phantom{$_\texttt{-1.3}$}80.5\textcolor{red}{$_\texttt{-1.3}$} & \phantom{$_\texttt{-1.6}$}85.7\textcolor{red}{$_\texttt{-1.6}$} & \phantom{$_\texttt{-14.5}$}60.3\textcolor{red}{$_\texttt{-14.5}$} \\
    \rowcolor{mygreen!60}
    \bf\method & \checkmark & \bf{83.1} & \bf{81.8} & \bf{87.3} & \bf{74.8} \\
    \bottomrule
    \end{tabular}}
    \caption{Zero-shot evaluation of our 2D-trained VLM on \bench, testing whether the model's representations are truly aligned. Our 2D model achieves reasonable accuracy without explicit 3D supervision.}
    \label{tab:abl_zeroshot}
    % \vspace{-0.15cm}
    % \vspace{-0.35cm}
\end{table}

\subsection{Analysis and Ablation Study}
\label{sec:exp:abl}
\myparagraph{Zero-shot Generalization.}
In this analysis, we aim to answer the question: Can a foundational 2D VLM trained exclusively on single-view image data perform zero-shot spatial reasoning on multi-view 3D scenes? To answer this, we evaluate its zero-shot performance on \bench covering Tall/Short, Big/Small, Height, and Distance categories. We exclude the width-related category because the width is defined differently in single-view and multi-view. In single-view images, width refers to the horizontal extent in the image plane~\citep{srgpt}, whereas in multi-view settings, it represents an object's maximum length or width. Table~\ref{tab:abl_zeroshot} shows our results, indicating that the single-view model performs highly competitively. This suggests that our unified representation design effectively transfers knowledge from single-view images, despite the challenge of the model never encountering multi-view data, scene-level position embeddings, or ground-truth spatial annotations.

\begin{table}[t]
    \small
    \centering
    \setlength{\tabcolsep}{5.5pt}
    \scalebox{0.97}{
    {\fontsize{8pt}{9pt}\selectfont
    \begin{tabular}{ccccccc}
    \toprule
    3D PE & 2D Pre-train & Scan2Cap & ScanQA & SQA3D & 3D Region & 3D Global \\
    \midrule
     & & 92.9 & 101.3 & 58.6 & 74.0 & 51.1\\
     & \ding{52} & 94.3 & 108.2 & 59.5 & 78.1 & 52.9 \\
    \ding{52} & & 92.7 & 102.9 & 59.1 & 75.3 & 51.2 \\
    \rowcolor{mygreen!60}
    \ding{52} & \ding{52} & \bf97.9 & \bf109.3 & \bf62.2 & \bf80.9 & \bf62.0 \\
    \bottomrule
    \end{tabular}}}
    % \vspace{-0.2cm}
    \caption{Ablation study on the impact of incorporating 3D positional embeddings (3D PE) and single-view pre-training. We report CIDEr for Scan2Cap and ScanQA, EM for SQA3D, and the average score for both the 3D region and global benchmarks.}
    \label{tab:abl-pe-and-pretrain}
    % \vspace{-0.5cm}
\end{table}

\myparagraph{3D Position Embedding and Single-view Pre-training.}
We conduct an ablation study to evaluate the impact of single-view pre-training and 3D positional embeddings on our model’s performance. We compare two model variants: one fine-tuned directly on multi-view data without positional embeddings and another with them. As shown in Table~\ref{tab:abl-pe-and-pretrain}, our results indicate that single-view pre-training significantly enhances performance on multi-view data by enabling the model to leverage prior spatial knowledge. In contrast, adding 3D positional embeddings without scaling provides only a marginal improvement. This highlights the necessity of scaling up to fully harness the power of positional representations for spatial reasoning.

\subsection{Applications}
\label{sec:exp:cut3r}

\begin{table}[!t]
    \scriptsize
    \centering
    % \vspace{-0.35cm}
    \setlength{\tabcolsep}{3.4pt}
    \begin{tabular}{lccccccccc}
    \toprule
     & 3D Source & C $\uparrow$ & B-1 $\uparrow$ & B-4 $\uparrow$ & M $\uparrow$ & R $\uparrow$ & EM $\uparrow$ \\
    \midrule
    Video3dLLM~\cite{video3dllm} & GT & 102.1 & 47.1 & 16.2 & 19.8 & 49.0 & 30.1 \\
    Video3dLLM~\cite{video3dllm} & Cut3R & 100.7 & 46.6 & 15.8 & 19.6 & 48.6 & 29.9 \\
    % Video3dllm & CUT3R & Norm. & 100.7 & 46.3 & 16.6 & 19.5 & 48.6 & 29.9 \\
    \midrule
    SR-3D & GT & 109.3 & 50.9 & 18.1 & 21.2 & 51.2 & 30.4 \\
    \rowcolor{mygreen!60}
    \bf SR-3D & \bf Cut3R & \bf109.3 & \bf50.9 & \bf18.1 & \bf21.2 & \bf51.2 & \bf30.2 \\
    \bottomrule
    \end{tabular}
    \caption{ScanQA results on both ground-truth point clouds and Cut3R-reconstructed point clouds.}
    \label{tab:cut3r}
\end{table}
Our method is flexible in two key ways.  
First, because \method is trained in a normalized 3D space, it naturally connects with existing 3D foundation models~\citep{cut3r,mast3r,wang2025pi,wang2025vggt} for pointmap estimation. The input is not restricted to 3D scans—\method can also operate on in-the-wild videos such as YouTube footage. To quantitatively validate this, we evaluate SR-3D on both ground-truth point clouds and Cut3R-reconstructed~\citep{cut3r} point clouds, comparing it with the baseline Video3dLLM~\cite{video3dllm} on ScanQA. As shown in Table~\ref{tab:cut3r}, \method maintains strong performance with Cut3R outputs, close to its ground-truth results, whereas the baseline exhibits a significant drop.  

Second, \method eliminates the need for costly 3D annotations or dense per-frame labeling. Instead, users can provide lightweight region inputs by simply drawing on a single frame, which the model then propagates for spatial reasoning across the video.  

Combining these two aspects, \method demonstrates robust spatial understanding from unconstrained video inputs without reliance on 3D scans or exhaustive annotations (Figure~\ref{fig:teaser}). These flexibilities open the door to a wide range of real-world applications, such as assisting robots in unstructured environments, analyzing large video collections, and supporting interactive spatial reasoning tasks.

% \method is trained in a normalized space, which makes it easy to connect with existing 3D foundation models for pointmap estimation. We provide examples in Figure~\ref{fig:teaser}, where the model takes in videos from YouTube and can perform both region-level and global spatial reasoning. \method is not limited to 3D scans, but is designed to support VLMs as spatial agents that help with decision-making and understanding object arrangements. 

% We also evaluate SR-3D on both ground-truth point clouds and Cut3R-reconstructed point clouds, comparing it with the baseline Video3dLLM~\cite{video3dllm} on ScanQA. The results in Table~\ref{tab:cut3r} show that \method maintains strong performance with Cut3R outputs, close to its ground-truth results, while the baseline drops.

\section{Conclusion}
\label{sec:conclusion}
We introduce SR-3D, a foundational VLM for 3D-aware spatial reasoning. By unifying single-view and multi-view data in a shared space, our approach leverages 2D priors from pretrained VLMs to tackle complex 3D tasks. Our tile-and-stitch method extracts high-resolution region features, enabling flexible region prompts across both settings. Experiments on 2D vision-language and 3D spatial benchmarks show state-of-the-art performance, validating SR-3D’s ability to unify and enhance spatial reasoning.

\clearpage
{
    \small
    \bibliographystyle{unsrtnat}
    \bibliography{main}
}

% Remove this before submission
\clearpage
\appendix

\setcounter{page}{1} % Reset page counter

% Enable ToC
\hypersetup{linkcolor=citecolor}

% Unnumbered Table of Contents
\section*{Appendix: Table of Contents}  % Use * to remove numbering
\startcontents[sections]
\printcontents[sections]{l}{1}{\setcounter{tocdepth}{3}}

% Reset section counter manually to ensure numbering starts from "A"
\setcounter{section}{0}
\clearpage
\section{More Quantitative Results on 3D General Benchmarks}

Following prior work, we report results using additional metrics for a more comprehensive evaluation. Table~\ref{sup:tab:scan2cap} presents results on Scan2Cap, Table~\ref{sup:tab:scanqa} on ScanQA, and Table~\ref{sup:tab:sqa3d} on SQA3D. Apart from our method, all other results are from Video-3D-LLM~\citep{video3dllm}.

\begin{table}[!ht]
\small
\centering
\setlength{\tabcolsep}{7pt}
    \scalebox{0.97}{
    {\fontsize{8pt}{9pt}\selectfont
\begin{tabular}{lcccccccc}
\toprule
 & Cider $\uparrow$ & Bleu-4 $\uparrow$ & Meteor $\uparrow$ & Rouge $\uparrow$\\
\midrule
Scan2Cap~\citep{chen2021scan2cap} & 39.1 & 23.3 & 22.0 & 44.5 \\
3DJCG~\citep{cai20223djcg}& 49.5 & 31.0 & 24.2 & 50.8 \\
D3Net~\citep{d3net}  & 62.6 & 35.7 & 25.7 & 53.9 \\
% Vote2Cap-DETR~\citep{vote2cap-detr} & 61.81 & 34.46 & 26.22 & 54.40 \\
3D-VisTA~\citep{zhu20233d}  & 66.9 & 34.0 & 27.1 & 54.3 \\
LL3DA~\citep{chen2024ll3da}  & 65.2 & 36.8 & 26.0 & 55.1 \\
LEO~\citep{huang2023embodied} & 68.4 & 36.9 & 27.7 & 57.8 \\
% Vote2Cap-DETR++~\citep{vote2cap-detr++} & 67.58 & {37.05} & 26.89 & 55.64 \\
ChatScene~\citep{chatscene} & 77.2 & 36.3 & 28.0 & 58.1 \\
LLaVA-3D~\citep{llava3d} & 79.2  &41.1 & 30.2 & 63.4 \\
Video-3D LLM~\citep{video3dllm} & 83.8 & 42.4 & 28.9 & 62.3 \\
% Video-3D LLM  (MC) & 80.00 & 40.18 & 28.49 & 61.68 \\
\rowcolor{mygreen!60}
\textbf{\method}& \bf97.9  & \bf44.7 & \bf31.5  & \bf67.3  \\
\bottomrule
\end{tabular}
}}
\vspace{-0.2cm}
\caption{{Full results on Scan2Cap~\citep{chen2021scan2cap} validation set.}} \label{sup:tab:scan2cap}
\end{table}

\begin{table}[!ht]
\small
\centering
\setlength{\tabcolsep}{2.5pt}
\scalebox{0.97}{
{\fontsize{8pt}{9pt}\selectfont
\begin{tabular}{lcccccccc}
\toprule
& EM & Bleu-1 $\uparrow$ & Bleu-2 $\uparrow$ & Bleu-3 $\uparrow$ & Bleu-4 $\uparrow$ & Rouge $\uparrow$ & Meteor $\uparrow$ & Cider $\uparrow$ \\
\midrule
ScanQA~\citep{azuma2022scanqa} & 21.1 & 30.2 & 20.4 & 15.1 & 10.1 & 33.3 & 13.1 & 64.9  \\
3D-VisTA \cite{zhu20233d}  & 22.4 & -- & -- & -- & 10.4 & 35.7 & 13.9 & 69.6 \\
Oryx-34B \cite{liu2024oryx}  & -- & 38.0 & 24.6 & -- & -- & 37.3 & 15.0 & 72.3 \\
LLaVA-Video-7B \cite{zhang2024video}  & -- & 39.7 & 26.6 & 9.3 & 3.2 & 44.6 & 17.7 & 88.7 \\
3D-LLM (Flamingo) \cite{hong20233d}  & 20.4 & 30.3 & 17.8 & 12.0 & 7.2 & 32.3 & 12.2 & 59.2 \\
3D-LLM (BLIP2-flant5)~\citep{hong20233d}  & 20.5 & 39.3 & 25.2 & 18.4 & 12.0 & 35.7 & 14.5 & 69.4  \\
Chat-3D \cite{wang2023chat}  & -- & 29.1 & -- & -- & 6.4 & 28.5 & 11.9 & 53.2 \\
NaviLLM \cite{zheng2024towards} & 23.0 & -- & -- & -- & 12.5 & 38.4 & 15.4 & 75.9 \\
LL3DA~\citep{chen2024ll3da}  & -- & -- & -- & -- & 13.5 & 37.3 & 15.9 & 76.8 \\
Scene-LLM~\citep{fu2024scene}  & 27.2 & 43.6 & 26.8 & 19.1 & 12.0 & 40.0 & 16.6 & 80.0 \\
LEO~\citep{huang2023embodied}  & -- & -- & -- & -- & 11.5 & 39.3 & 16.2 & 80.0 \\
Grounded 3D-LLM~\citep{chen2024grounded3dllm} & -- & -- & -- & -- & 13.4 & -- & -- & 72.7\\
ChatScene \cite{chatscene}  & 21.6 & 43.2 & 29.1 & 20.6 & 14.3 & 41.6 & 18.0 & 87.7 \\
LLaVA-3D \cite{llava3d}  & 27.0 & -- & -- & -- & 14.5 & 50.1 & 20.7 &  91.7 \\
Video-3D LLM~\citep{zhang2024video} & 30.1 & 47.1 & 31.7 & 22.8 & 16.2 & 49.0 & 19.8 & 102.1 \\
\rowcolor{mygreen!60}
\textbf{\method}  & \bf30.4 & \bf50.9 & \bf34.3 & \bf25.1 & \bf18.1 & \bf51.2 & \bf21.1 & \bf109.3\\
\bottomrule
\end{tabular}}}
\vspace{-0.2cm}
\caption{Full results on ScanQA \cite{azuma2022scanqa} validation set.
} \label{sup:tab:scanqa}
\end{table}

\begin{table}[!ht]
\small
\centering
\setlength{\tabcolsep}{4pt}
    \scalebox{0.97}{
    {\fontsize{8pt}{9pt}\selectfont
\begin{tabular}{lccccccc}
\toprule
 & What & Is & How & Can & Which & Others & Avg. \\
\midrule
SQA3D~\citep{ma2024sqa3d} & 31.6 & 63.8 & 46.0 & 69.5 & 43.9 & 45.3 & 46.6 \\
3D-VisTA~\citep{zhu20233d} & 34.8 & 63.3 & 45.4 & 69.8 & 47.2 & 48.1 & 48.5 \\
LLaVA-Video\cite{zhang2024video} & 42.7 & 56.3 & 47.5 & 55.3 & 50.1 & 47.2 & 48.5 \\
Scene-LLM~\citep{fu2024scene} & 40.9 & 69.1 & 45.0 & 70.8 & 47.2 & 52.3 & 54.2 \\
LEO~\citep{huang2023embodied} & -- & -- & -- & -- & -- & -- & 50.0 \\
ChatScene~\citep{} & 45.4 & 67.0 & 52.0 & 69.5 & 49.9 & 55.0 & 54.6 \\
LLaVA-3D~\citep{llava3d} & -- & -- & -- & -- & -- & -- & 55.6 \\
Video-3D LLM~\citep{video3dllm} & 51.1 & 72.4 & 55.5 & 69.8 & 51.3 & 56.0 & 58.6 \\
\rowcolor{mygreen!60}
\textbf{\method} & \bf55.0 & \bf76.4 & \bf59.8 & \bf71.6 & \bf54.7 & \bf61.1 & \bf62.2 \\
\bottomrule
\end{tabular}
}}
\vspace{-0.2cm}
\caption{Full results on SQA3D~\citep{ma2024sqa3d} testing set.}
\label{sup:tab:sqa3d}
\end{table}

\section{More Quantitative Results on VSI-Bench}
\begin{figure*}[!ht]
    \centering
\includegraphics[width=1.0\linewidth]{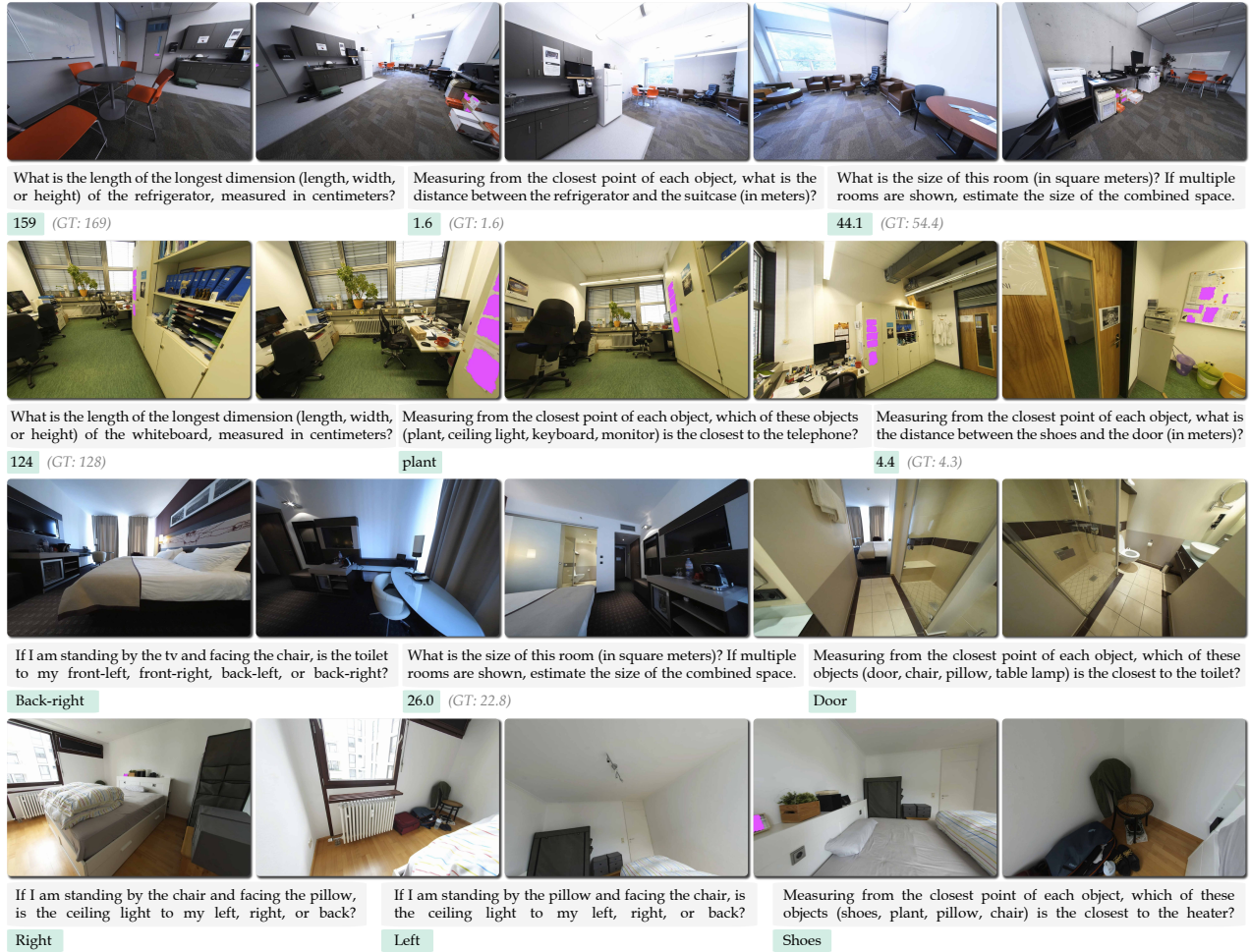}
\vspace{-0.2cm}
    \caption{More results on VSI-Bench~\citep{yang2024thinking}. We highlight SR-3D ’s outputs and include ground-truth values for numerical answers.}
    \label{fig:sup:more-vsi}
\end{figure*}

We report additional visual results on VSI-Bench, primarily using scenes from ScanNet\textsuperscript{++}. ScanNet\textsuperscript{++} is not included in EmbodiedScan's annotations, making it a distinct and challenging dataset for evaluation. Compared to ScanNet, ScanNet\textsuperscript{++} offers higher fidelity and greater diversity in indoor environments. Moreover, its 3D annotations are only coarsely aligned to match walls and floors to the axis. Despite these challenges, as shown in Figure~\ref{fig:sup:more-vsi}, our method demonstrates superior capabilities in determining relative direction, highlighting its robustness in real-world tasks.

\section{More Ablation Study}
We present the complete ablation study results on 2D single-view pre-training and 3D positional encoding without pre-training, evaluating their influence on model performance. The detailed results are shown in Table~\ref{tab:sup:full-abl-general} and Table~\ref{tab:sup:full-abl-spatial}, respectively.

Overall, the fully-trained model consistently outperforms baseline models on 3D general QA benchmarks, demonstrating the benefits of leveraging both 2D and 3D spatial information. However, in the 3D spatial-focused dataset, we observe a slight drop in the Wide and Big category, likely due to differences in how width is defined in 2D versus 3D, as discussed in the main paper. 

Additionally, we find that removing pre-training leads to a substantial drop in performance for more complex reasoning tasks, particularly in the multi-choice complex category, where the model struggles without prior exposure to large-scale 2D pre-training. These results highlight the importance of both spatial-aware representation learning and strong pre-training strategies in enhancing 3D reasoning capabilities.

\begin{table}[!ht]
\small
\centering
\setlength{\tabcolsep}{6pt}
\scalebox{0.95}{
{\fontsize{8pt}{9pt}\selectfont
\begin{tabular}{llcccccccccc}
\toprule
 & & \multicolumn{4}{c}{Scan2Cap} & \multicolumn{4}{c}{ScanQA} & \multicolumn{2}{c}{SQA3D} \\
\cmidrule(lr){3-6} \cmidrule(lr){7-10} \cmidrule(lr){11-12}
PE & PT & Bleu-4 $\uparrow$ & Rouge$\uparrow$ & Cider$\uparrow$ & Meteor$\uparrow$ & Bleu-4 $\uparrow$ & Rouge$\uparrow$ & Cider$\uparrow$ & Meteor$\uparrow$ & EM $\uparrow$ & EM $\uparrow$ \\
\midrule
& & 44.2 & 67.3 & 92.9 & 31.1 & 16.0 & 48.9 & 101.3 & 19.8 & 28.8 & 58.6 \\
\ding{52} & & 44.0 & 67.3 & 92.7 & 31.0 & 17.4 & 48.8 & 102.9 & 20.0 & 29.1 & 59.1 \\
\rowcolor{mygreen!60}
\ding{52} & \ding{52} & \bf44.7 & \bf67.3 & \bf97.9 & \bf31.5 & \bf18.1 & \bf51.2 & \bf109.3 & \bf21.2 & \bf30.4 & \bf62.2 \\
\bottomrule
\end{tabular}}}
\vspace{-0.2cm}
\caption{Ablation study full results on Scan2Cap, ScanQA, and SQA3D benchmarks.}
\label{tab:sup:full-abl-general}
\end{table}

\begin{table*}[!ht]
\small
\centering
\setlength{\tabcolsep}{4pt}
\scalebox{0.97}{
{\fontsize{8pt}{9pt}\selectfont
\begin{tabular}{lccccccccc}
\toprule
Category & Thin-Wide & Tall-Short & Big-Small & Multi-Simple & Multi-Complex & Width Data & Distance Data & Height Data & Total Length \\
\midrule
Count    & 219      & 231        & 231       & 117          & 500           & 496        & 242           & 464        & 2500 \\
\bottomrule
\end{tabular}}}
\vspace{-0.2cm}
\caption{Statistical analysis of our \bench, showing the distribution of different spatial attributes.}
\label{tab:bench_stats}
\end{table*}

\begin{table*}[!ht]
    \footnotesize\centering
    \begin{tabular}{rp{360pt}}
        \toprule
        \multicolumn{2}{l}{\textit{\textbf{2D Data}}} \\
        \midrule
        Hybrid     & ShareGPT4V-SFT, Molmo, The Cauldron, Cambrian, LLaVA-OneVision \\
        \midrule
        Captioning & MSR-VTT, Image Paragraph Captioning, ShareGPT4V-100K \\
        \midrule
        Reasoning  & CLEVR, NLVR, VisualMRC \\
        \midrule
        Document & DocVQA, UniChart-SFT, ChartQA \\ 
        \midrule
        OCR & TextCaps, OCRVQA, ST-VQA, POIE, SORIE, SynthDoG-en, TextOCR-GPT4V, ArxivQA, LLaVAR \\
        \midrule
        General VQA & ScienceQA, VQAv2, ViQuAE, Visual Dialog, GQA, Geo170K, LRV-Instruction, RefCOCO, GeoQA, OK-VQA, TabMVP, EstVQA \\
        \midrule
        Diagram \& Dialogue & DVQA, AI2D, Shikra, UniMM-Chat \\
        \midrule
        Instruction & LRV-Instruction, SVIT, MMC-Instruction, MM-Instruction \\
        \midrule
        Text-only & FLAN-1M, MathInstruct, Dolly, GSM8K-ScRel-SFT \\
        \midrule
        Knowledge & WordART, WIT, STEM-QA \\
        \midrule
        Medical & PathVQA, Slake, MedVQA \\
        \midrule
        Region & RegionGPT \\
        \midrule
        Spatial & SpatialRGPT \\
        \midrule
       \multicolumn{2}{l}{\textit{\textbf{3D Data}}} \\
        \midrule
        General & ScanQA, SQA3D, Scan2Cap \\
        \midrule
        Spatial & EmbodiedScan \\
        \bottomrule
    \end{tabular}
    \vspace{-0.2cm}
    \caption{Data recipe for training 2D foundational VLM and 3D fine-tuning.}
    \label{tab:recipe:data}
\end{table*}

\begin{table}[!ht]
\small
\centering
\setlength{\tabcolsep}{5pt}
\scalebox{0.95}{
{\fontsize{8pt}{9pt}\selectfont
\begin{tabular}{llcccccccccc}
\toprule
 & & \multicolumn{6}{c}{3D Region} & \multicolumn{4}{c}{3D Global} \\
\cmidrule(lr){3-8} \cmidrule(lr){9-12}
PE & PT & Wide & Tall & Big & M. Sim. & M. Cpx. & Avg. & Width & Height & Dist. & Avg. \\
\midrule
& & \bf77.6 & 80.5 & \bf82.6 & 71.7 & 55.8 & 73.6 & 85.8 & 84.4 & 53.7 & 74.4 \\
\ding{52} & & \bf77.6 & \bf83.1 & 80.5 & 70.9 & 59.0 & 74.2 & 85.5 & 85.7 & 60.3 & 77.2 \\
\rowcolor{mygreen!60}
\ding{52} & \ding{52}& 76.3 & \bf83.1 & 81.8 & \bf80.3 & \bf76.0 & \bf79.5 & \bf87.7 & \bf87.3 & \bf74.8 & \bf83.3 \\
\bottomrule
\end{tabular}}}
\vspace{-0.2cm}
\caption{Ablation study full results.}
\label{tab:sup:full-abl-spatial}
\end{table}

\section{Statistics of \bench}
Our benchmark follows template designs from prior works on spatial reasoning in vision-language models, including SpatialRGPT and SpatialVLM. To further increase the complexity and diversity of spatial reasoning tasks, we incorporate situated annotations from the EmbodiedScan dataset, ensuring a more realistic and challenging evaluation setting. Specifically, our dataset includes a range of spatial relationships, from basic geometric comparisons such as thin-wide, tall-short, and big-small, to more complex multi-object interactions categorized as multi-simple and multi-complex. Additionally, we introduce explicit width, distance, and height annotations to facilitate fine-grained spatial understanding. With a total of 2,500 samples, our benchmark provides a comprehensive evaluation for assessing the region-level spatial reasoning capabilities of vision-language models in realistic scenarios.

\section{Implementation Details of \method}
We use PaliGemma as our visual backbone with an input size of 448 and a patch size of 14, paired with a Qwen-2-7B LLM backbone. For training the foundational 2D VLM, we follow prior work and set the maximum tiles per image to 12. For the multi-view VLM, we use a frame size of 32 with a uniform sampling strategy to ensure a fair comparison with previous methods. For training the 2D VLM, we adopt a learning rate of 5e-5 with cosine decay and gradient clipping enabled. The same hyperparameters are used for fine-tuning the 3D VLM, except for a reduced batch size due to the increased token length. The data recipe for both training stages is detailed in Table~\ref{tab:recipe:data}. We train on a subset of 2D data, excluding spatial and region-related datasets, to preserve the original vision-language capabilities while incorporating a diverse source.

\section{Limitations}

\myparagraph{Orientations}  
Although our method shows promising results, it remains challenging for current vision-language models to accurately perceive and interpret spatial questions related to object orientation. This challenge arises due to the difficulty of scaling up data. We leave this as future work.

\myparagraph{Dynamic Videos}  
Our method is designed for multi-view static data, whereas real-world scenarios often involve dynamic environments. Incorporating positional embeddings to handle both static and dynamic inputs is non-trivial. Future work should explore methods to address this limitation.

\myparagraph{OCR Tasks}  
In the main paper, Table~\ref{tab:2d_general}, we report the performance of our 2D foundation model on general benchmarks. While our model maintains comparable performance to the base model, demonstrating improved spatial understanding without significant trade-offs, we observe a consistent slight drop in OCR-related tasks. A potential solution is to incorporate more OCR-related tasks into the training data pipeline.

\myparagraph{Unified Checkpoint}  
While our unified architecture and representation provide a foundation for both single- and multi-view 3D-aware VLMs, we leave it to future work to investigate how to effectively combine the two models. This could be achieved either by introducing an agentic flow between single- and multi-view models or by directly training a single model across both settings, which may further improve generalization and efficiency.

\end{document}